\documentclass[journal]{IEEEtran}

\ifCLASSINFOpdf
\else
\fi

\usepackage{graphics}
\usepackage{graphicx}
\usepackage{epstopdf}
\usepackage{textcomp}
\usepackage{longtable}
\usepackage[table]{xcolor}
\usepackage{amsmath, amsthm, amssymb}
\definecolor{light-gray}{gray}{0.85}
\usepackage{multirow}
\usepackage[square, numbers, comma, sort&compress]{natbib}

\begin{document}
%
\title{People Counting in High Density Crowds from Still Images}
%
%
%


\author{Ankan~Bansal, and
        K S Venkatesh
\thanks{A. Bansal (corresponding author) and K. S. Venkatesh are with the Department 
        of Electrical Engineering, Indian Institute of Technology, Kanpur, 208016, India (e-mail: ankan@iitk.ac.in).}}
\maketitle

\begin{abstract}
We present a method of estimating the number of people in high density crowds from still images. The method estimates counts by fusing information from multiple sources. Most of the existing work on crowd counting deals with very small crowds (tens of individuals) and use temporal information from videos. Our method uses only still images to estimate the counts in high density images (hundreds to thousands of individuals). At this scale, we cannot rely on only one set of features for count estimation. We, therefore, use multiple sources, \emph{viz}. interest points (SIFT), Fourier analysis, wavelet decomposition, GLCM features and low confidence head detections, to estimate the counts. Each of these sources gives a separate estimate of the count along with confidences and other statistical measures which are then combined to obtain the final estimate. We test our method on an existing dataset of fifty images containing over 64000 individuals. Further, we added another fifty annotated images of crowds and tested on the complete dataset of hundred images containing over 87000 individuals. The counts per image range from 81 to 4633. We report the performance in terms of mean absolute error, which is a measure of accuracy of the method, and mean normalised absolute error, which is a measure of the robustness.
\end{abstract}

%
\IEEEpeerreviewmaketitle

\section{Introduction}
\label{sec:intro}

Crowd counting is one of the first and foremost parts of crowd management. It has several real-world applications like crowd management, safety control and urban planning, monitoring crowds for surveillance, modelling crowds for animation and crowd simulation. Crowd size may also be an indicator of comfort level in public spaces or of an imminent stampede.

\begin{figure}
\begin{tabular}{cc}
\includegraphics[width=3.9cm,height=2.6cm]{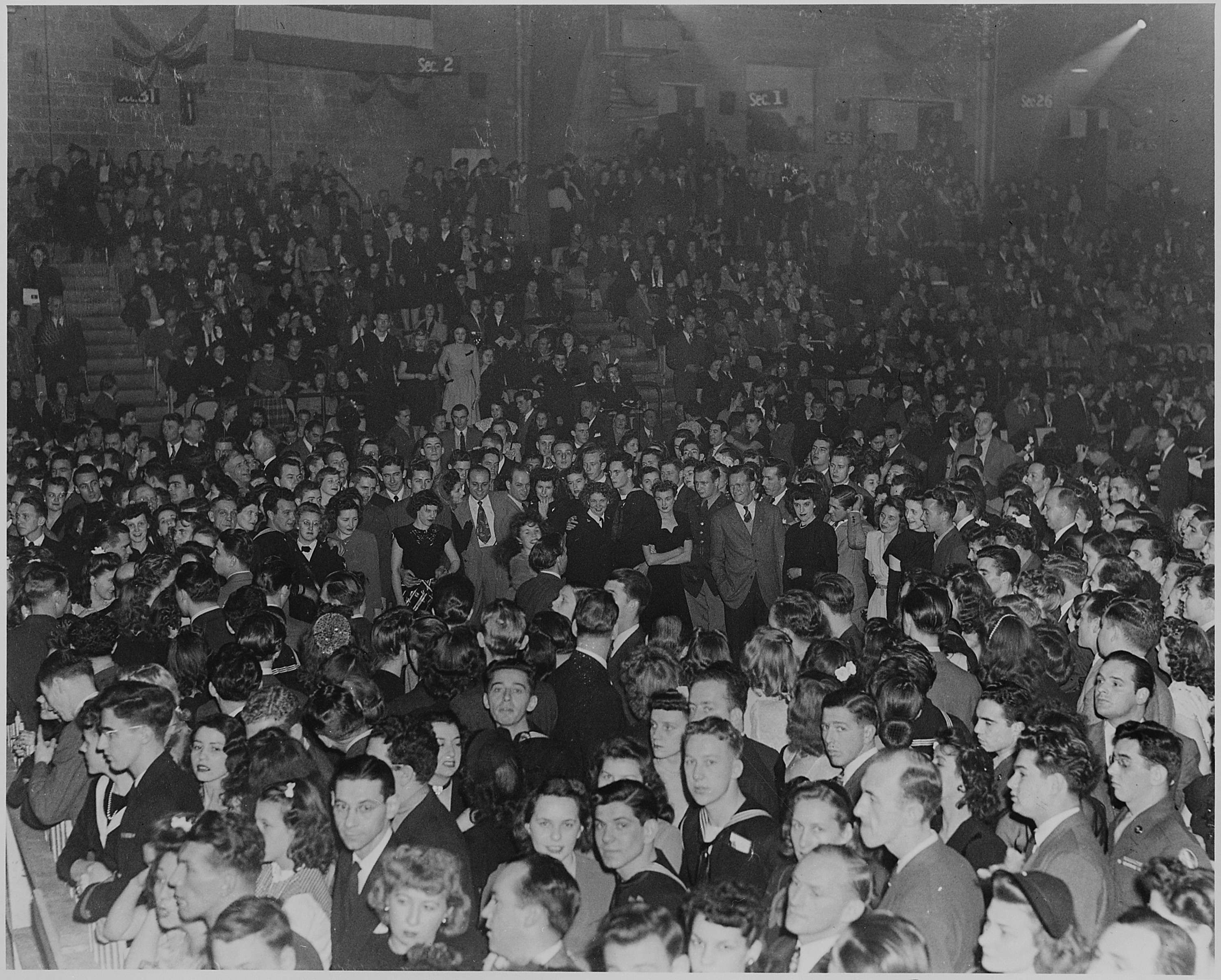}&
\includegraphics[width=3.9cm,height=2.6cm]{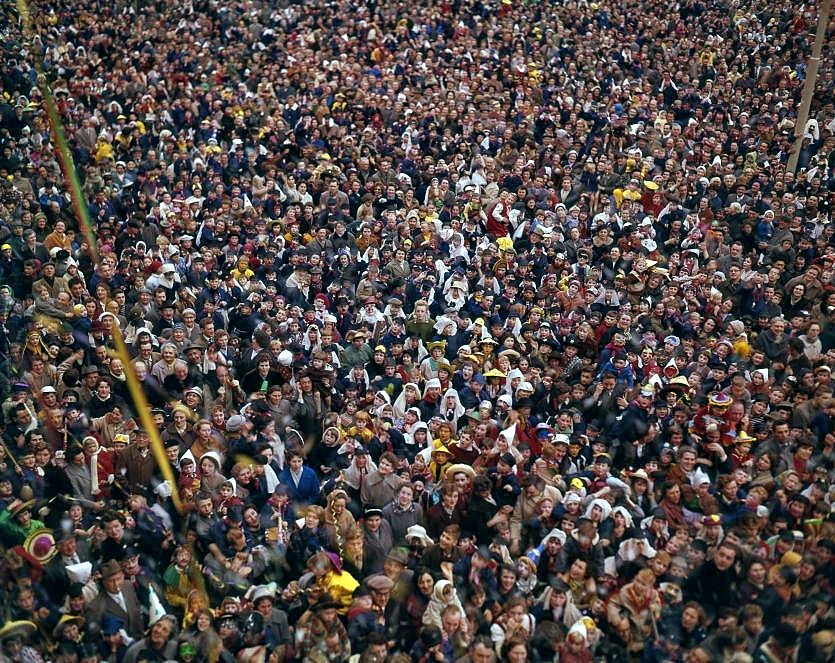}\\
\includegraphics[width=3.9cm,height=2.6cm]{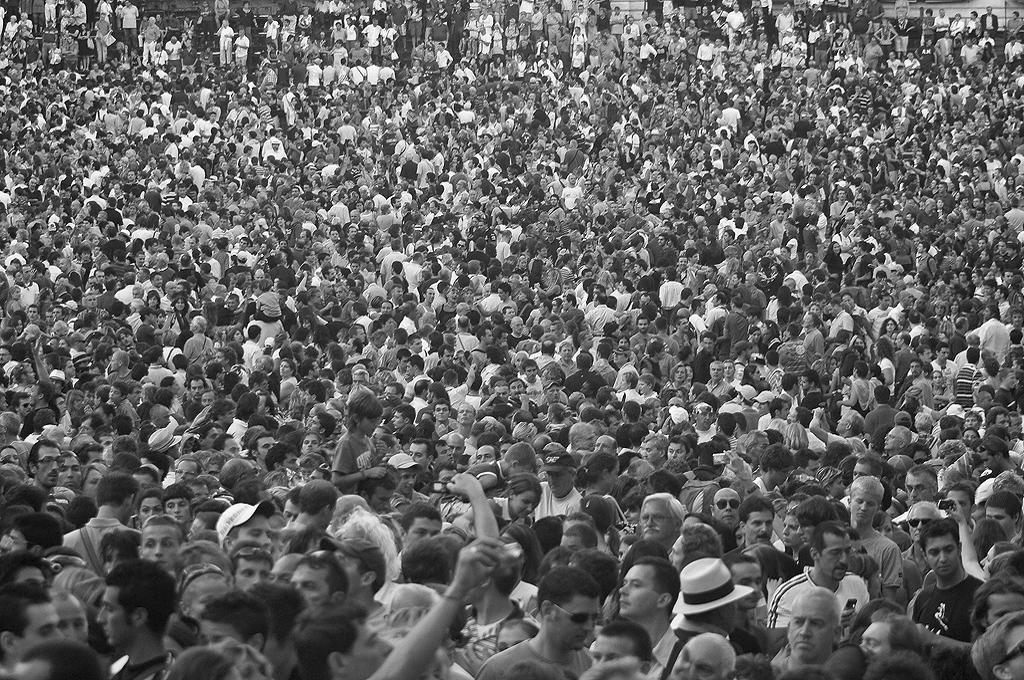}&
\includegraphics[width=3.9cm,height=2.6cm]{53.jpg}
\end{tabular}
\caption{Examples of high density crowds. On average, each image in our crowd counting dataset contains 870 people with a minimum of 81 and maximum of 4633.}
\label{fig:teaser}
\end{figure}

Many automated systems for density and count estimation have been proposed. However, many of these techniques suffer from some limitations: inability to handle large crowds (hundreds or thousands of people); reliance on temporal constraints in crowd videos; reliance on detecting, tracking and analysing individual persons in crowds. Another important limitation that some of these methods suffer from is the requirement of installed infrastructure on the site. 

Most existing people counting methods can be divided into three categories: (1) pixel-based analysis; (2) texture-based analysis; and (3) object-level analysis. Pixel based methods employ very local features such as edge information or individual pixel analysis to obtain counts~\cite{Chan2008a,Kong2005,Lempitsky2010,Ryan2009,Chen2012,Li2008,Chan2012,Zhao2008}. Since they use very local features, these methods are mostly focussed on density estimation rather than the count. Texture based methods rely on texture modelling through the analysis of image patches~\cite{Arandjelovic2008,Chen2012,Ma2010,Chan2009,Chan2008a,Chan2012,Marana1998,Marana1999,Idrees2013}. Some texture analysis methods that have been suggested include grey-level co-occurrence matrix, Fourier analysis and fractal dimension. Object level analysis methods try to locate individual persons in a scene~\cite{Brostow2006,Ge2009,Li2008,Rabaud,Rodriguez2011,Zhao2008,Felzenszwalb2008,Felzenszwalb2010,Idrees2013}. But these methods work best only for very low density crowds. For high density crowds the evidence for the presence of a single person is scarce. Even for low density crowds, partial occlusions, variations in clothing and pose, perspective effects, cluttered background and other complexities negatively impact the performance of object-level methods.

Brostow and Cipolla~\cite{Brostow2006} and Rabaud and Belongie~\cite{Rabaud} count moving people in videos by estimating contiguous regions of coherent motion. Addressing concerns about preservation of privacy in tracking people for counting, Chen \emph{et al.}~propose texture based method for counting~\cite{Chan2008a}. They use a mixture of dynamic textures to segment the crowd into components of homogeneous motion and then use a set of simple holistic features to find the correspondence between features and the number of individuals per segment.

Some works estimate the relationship between low-level features and the density or count by training regression models. Some of these methods are global, which learn a single regression function for the entire image/video~\cite{Chan2008a,Kong2005,Ryan2009,Cho1999}. But these methods make an implicit assumption that the density is the same over the image, which is not valid for most images due to perspective effects, viewpoint changes etc. Some regression methods can be local which divide the image into cells and perform regression for each cell~\cite{Ma2010,Lempitsky2010,Idrees2013}. These methods deal with the problems associated with global regression methods efficiently. An alternate multi-output regression model was proposed by Chen \emph{et al.}~in~\cite{Chen2012}.

The aim of this paper is to develop an effective texture-based method to solve the problem of counting the number of people in extremely dense crowds. Our goal is to arrive at a method that works well for dense crowds but at the same time is robust to variations in density. For very dense crowds, a single feature or detection method alone can not provide an accurate count due to low resolutions, occlusions, foreshortening and perspective. We build upon the work of Idrees \emph{et al.}~\cite{Idrees2013} and propose a model that combines sources of complementary information extracted from the images. Dense crowds can be thought of as a texture and this texture corresponds to a harmonic pattern at fine scales. Texture analysis methods have shown promising results for crowd counting/density estimation~\cite{Chan2009,Chan2008a,Ma2010}. 

Appearance based features like SIFT descriptors are also useful to estimate the texture elements. SIFT features have been shown to be successful for crowd detection in \cite{Arandjelovic2008}. Idrees \emph{et al.}~also used SIFT as one of the features for estimating crowd counts.

The main contribution of this work is the use of multiple texture analysis sources to estimate the counts for dense crowds. We employ Fourier analysis, GLCM features and wavelet transform to analyse the texture information. As far as we know, wavelet features have not been used for crowd counting or density estimation before. Along with these, we use head-detections and SIFT descriptors for our framework. The combination of multiple information sources provides robustness and accuracy to the counting process.

Existing methods suffer from severe scalability issues. Most existing methods have been tested on low to medium density crowds, e.g. , UCSD dataset~\cite{Chan2008a} (11 - 46 people per frame), Mall dataset~\cite{Chen2012} (13 - 53 people per frame) and PETS dataset~\cite{Ferryman2010} (3 - 40 people per frame). In contrast, we show the performance of our method on the UCF crowd counting dataset~\cite{Idrees2013} of 50 images containing between 96 and 4633 people per image. Further, we complement the dataset with 50 more images to expand it to 100 images and demonstrate the robustness and accuracy of our method on this new combined dataset.

The remainder of the paper is organised as follows. We present our methodology in section~\ref{sec:method}, provide experimental validation and results in section~\ref{sec:exp&res} and finally discuss and draw some conclusions in section~\ref{conclusion}.

\section{Methodology}
\label{sec:method}
\begin{figure}
\begin{center}
\begin{tabular}{c}
\includegraphics[scale=0.15]{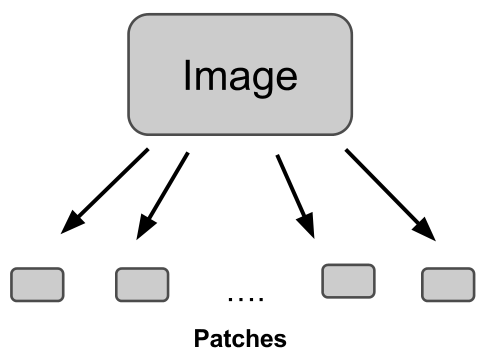}\\
(a)\\
\\
\includegraphics[scale=0.17]{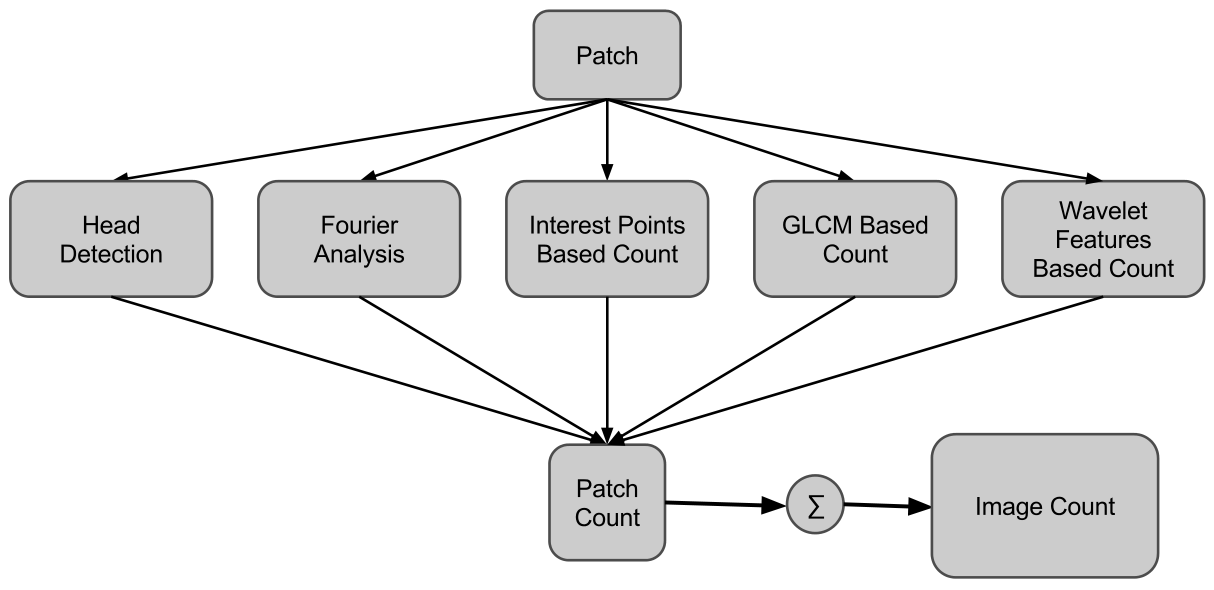}\\
(b)\\
\end{tabular}
\end{center}
\caption{A flow chart illustrating the methodology adopted in this paper. (a) The image is first divided into small cells in a grid. (b) Count is estimated for each cell by fusing estimates from five different methods. The final count estimate for the image is the sum of all cell counts.}
\label{fig:flowchart}
\end{figure}

Our goal is to provide an estimate of the total person count in an image. Figure~\ref{fig:flowchart} gives an overview of our framework. We first partition the image into small cells in a grid. We obtain the count estimate from each cell to counter the variations in density of the crowd over the image. The final output is the sum of all cell counts.

\subsection{Counting in a cell}
For a given cell $P$, we estimate the counts and confidences from five different sources. These are combined to obtain a final estimate of the person count for that cell. (Note: In this section we will use `cell'/ `patch' to mean the same thing.)

\subsubsection{\textbf{Interest-points based count}}
Arandjelovi\'{c}~\cite{Arandjelovic2008} used a statistical model based on quantised SIFT features to segment an image into crowd and non-crowd regions. Subsequently, Idrees \emph{et al.}~\cite{Idrees2013} used interest points to estimate counts and to get a confidence score of whether the cell represents a crowd. We follow this idea to calculate the counts. Given a training set, we obtain SIFT features and cluster then into a codebook of size $K$. We use the sparse SIFT features to train a Support Vector Regression model using the counts at each patch from ground truth and then use the trained model to obtain counts for new images patches. We calculate the SIFT features using the VL-FEAT library~\cite{Vedaldi08vlfeat}.

Due to the sparse nature of SIFT features, the probability of observing $k_i$ instances of the $i$-th SIFT word can be modelled as a Poisson distribution. Suppose, for a cell containing crowd, the expected number of detections of the $i$-th SIFT word is $\lambda_i^{+}$. Then: 

\begin{equation}
\label{eqn:poisson_pos}
p(k_i|crowd) = \frac{e^{-\lambda_i^+}[\lambda_i^+]^{k_i}}{k_i!}
\end{equation}

Similarly for a non-crowd cell ($\lambda_i^{-}$):

\begin{equation}
\label{eqn:poisson_neg}
p(k_i|\neg crowd) = \frac{e^{-\lambda_i^-}[\lambda_i^-]^{k_i}}{k_i!}
\end{equation}

Assuming independence between counts of any two SIFT words in a cell:

\begin{equation}
\label{eqn:indep}
p(k_i , k_j|crowd) = p(k_i|crowd)p(k_j|crowd)
\end{equation}

The log of the likelihood ratio of crowd and non-crowd patches is:

\begin{equation}
\label{eqn:likeratio}
\begin{aligned}
\mu &= \log p(k_1,k_2,\dots,k_K|crowd) - \log p(k_1,k_2,\dots,k_K|\neg crowd) \\
& = \sum_{i=1}^{K}[\lambda_i^- - \lambda_i^+ + k_i(\log \lambda_i^+ - \log \lambda_i^-)]
\end{aligned}
\end{equation}

$\mu$ can be interpreted as the confidence about the presence of crowd in an image cell.

\hspace{0.5cm}

\subsubsection{\textbf{Counts from texture analysis methods}}
Crowds are repetitive in nature since all humans appear similar from a distance. We employ three different texture analysis methods which separately give an estimate of the count which will be used later to give a final estimate of the count in the cell.
\hspace{0.3cm}
\begin{itemize}
\item Fourier analysis\\
Fourier analysis can be used to capture the repetitions in crowds. Since we are dealing with small cells and not the complete image, we can safely assume that the crowd density in a cell is uniform. In this case, the Fourier transform, $f(\omega)$, will show the repeated occurrence of people as peaks. 

For a given cell, $P$, in an image, we calculate the gradient, $\nabla(P)$, and apply a low pass filter to remove high frequency components. Then we apply inverse Fourier transform to obtain the reconstructed image patch, $P_r$. The local maxima in the reconstructed image give an estimate of the total person count in that cell. Figure~\ref{fig:fourier} shows some images with the local maxima obtained by this method marked. We also calculate the several other statistical measures, such as entropy, mean, variance, skewness and kurtosis for both $P_r$ and the difference $|\nabla(P) - P_r|$. We use the count and these measures as input for the next step (section \ref{subsec:total}).

\begin{figure}
\begin{tabular}{cc}
\includegraphics[width=3.9cm,height=2.6cm]{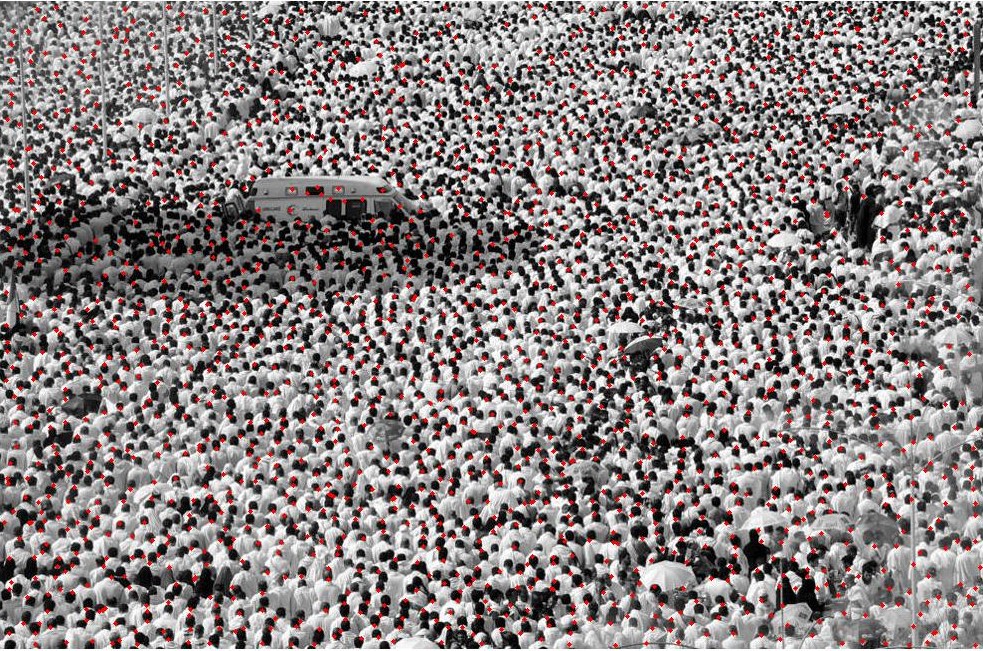}&
\includegraphics[width=3.9cm,height=2.6cm]{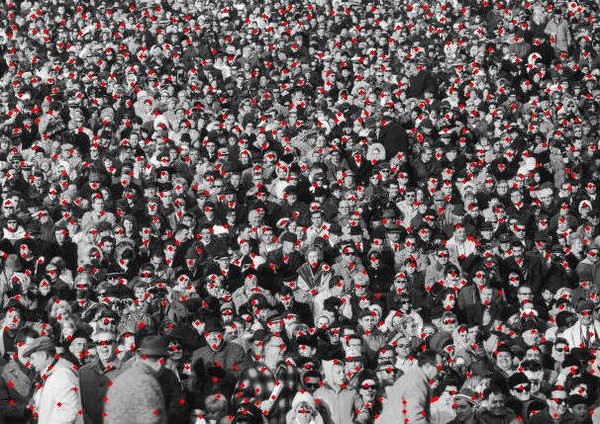}\\
(a)&(b)\\
\includegraphics[width=3.9cm,height=2.6cm]{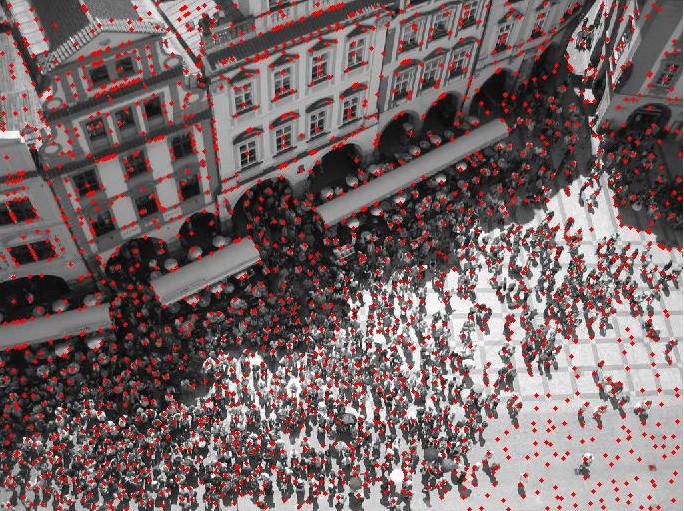}&
\includegraphics[width=3.9cm,height=2.6cm]{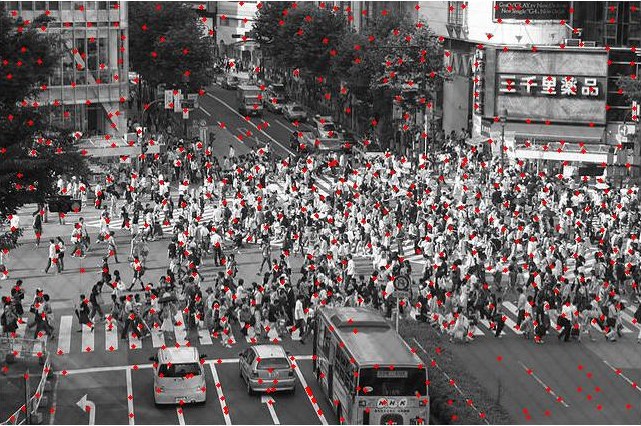}\\
(c)&(d)
\end{tabular}
\caption{Local maxima (red dots) for some images obtained by Fourier analysis. (a), (b) We notice that the maxima correspond very well with heads in dense crowds. (c), (d) However, Fourier analysis is crowd blind and cannot detect the actual presence of a crowd.}
\label{fig:fourier}
\end{figure}

\hspace{0.3cm}
\item GLCM features\\
Marana \emph{et al.} used texture features based on the gray-level co-occurrence matrix (GLCM) to classify image patches into categories based on density~\cite{Marana1998}. Other people have also used GLCM features for density/count estimation~\cite{Chan2008,Chan2009,Ma2010}. We adopt similar features to estimate the number of people. We quantise the image and calculate the joint-conditional probability density function, $f(i,j|d,\theta)$, with distance, $d = 1$, and angles, $\theta \in \{0\mbox{\textdegree}, 45\mbox{\textdegree}, 90\mbox{\textdegree}, 135\mbox{\textdegree}\}$. We calculate the following texture features:
Dissimilarity: $D(d,\theta) = \sum_{i,j}f(i,j|d,\theta)|i-j|$; Homogeneity: $H(d,\theta) = \sum_{i,j}\frac{f(i,j|d,\theta)}{1 + (i-j)^2}$; Energy: $E(d,\theta) = \sum_{i,j}f(i,j|d,\theta)^2$; Entropy: $P(d,\theta) = -\sum_{i,j}f(i,j|d,\theta)\log f(i,j|d,\theta)$.
So we obtain 16 (four for each $\theta$) features for each image cell. We then train a support vector regression model using these features and ground truths from cells from the training images. We pass the count estimate and statistical measures such as variance, skewness and kurtosis of the GLCM matrices as features to the next stage.

\hspace{0.3cm}
\item Wavelet decomposition\\
We use the multi-resolution properties of the two-dimensional wavelet transform to extract features for the counting framework. 

Given a cell, $P$, we calculate the three-step pyramid-structured wavelet transform and obtain the 10 lower resolution sub-images. We then calculate the energies contained in each of them as:
$e = \frac{1}{M\times N}\sum_{i=1}^{M}\sum_{j=1}^{N}|I(i,j)|$, 
where $I$ is a sub-image with resolution $M\times N$. Thus, we obtain a ten-dimensional feature vector for each image cell. Texture energies are distributed differently for different texture patterns. Thus the energy features calculated above can be used for discriminating crowds and estimating crowd counts. We train a support vector regression with these feature vectors using the ground truth counts from the training data as outputs. We also calculate statistical measure such as variance, skewness and kurtosis of the 10 lower resolution sub-images and pass these measures along with the count to the next step.

\end{itemize}

\hspace{0.5cm}
\subsubsection{\textbf{Count from head detections}}
Detecting humans is not possible in dense crowds due to severe occlusions. Only the heads may be visible at this scale. So we estimate the count by detecting heads in the image. We used a Deformable Part Model~\cite{Felzenszwalb2008} trained on the INRIA Person dataset and applied only the filter corresponding to heads with a low threshold. This is because heads are often very small and partially occluded in such images.

\begin{figure}
\begin{center}
\begin{tabular}{cc}
\includegraphics[width=3.9cm,height=2.6cm]{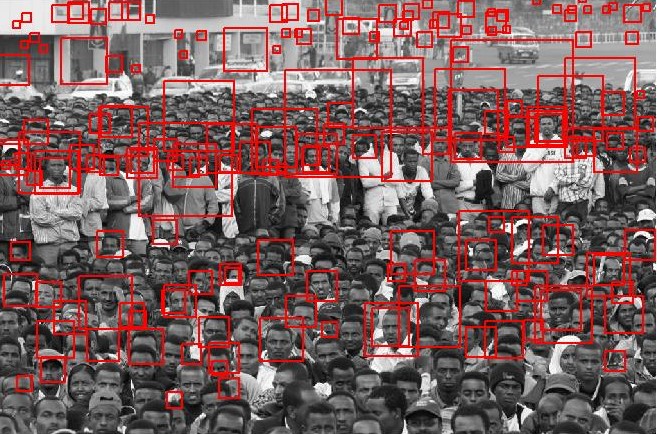}&
\includegraphics[width=3.9cm,height=2.6cm]{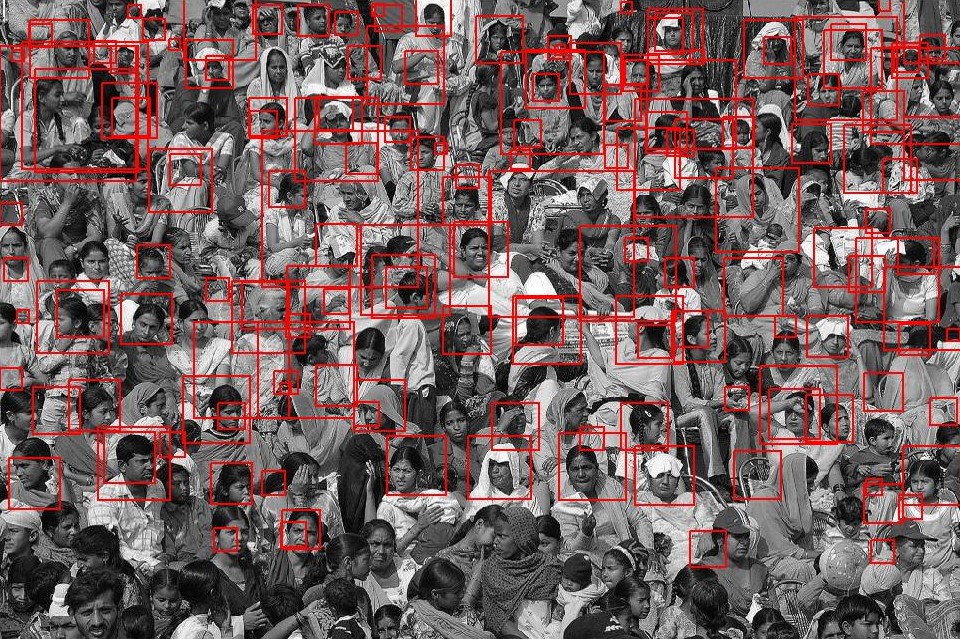}\\
\includegraphics[width=3.9cm,height=2.6cm]{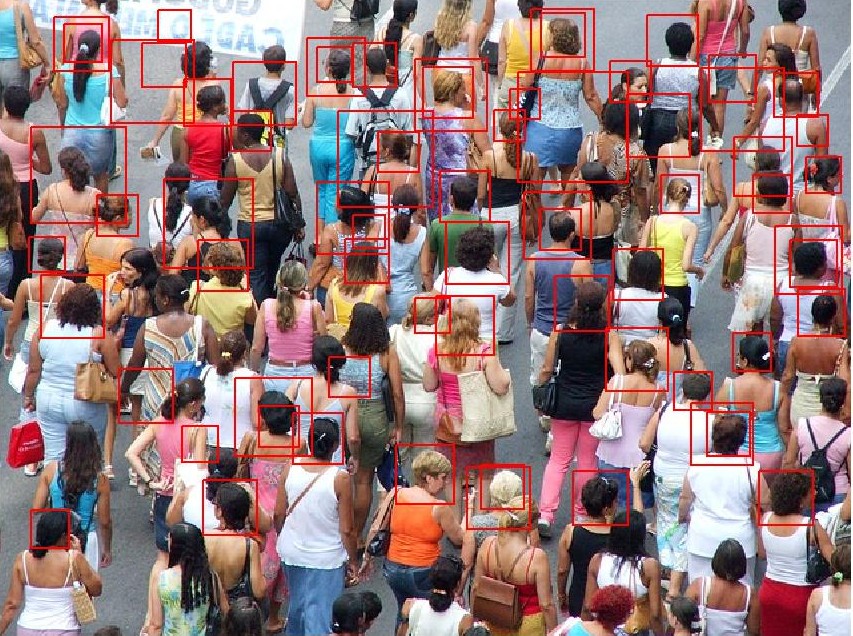}&
\includegraphics[width=3.9cm,height=2.6cm]{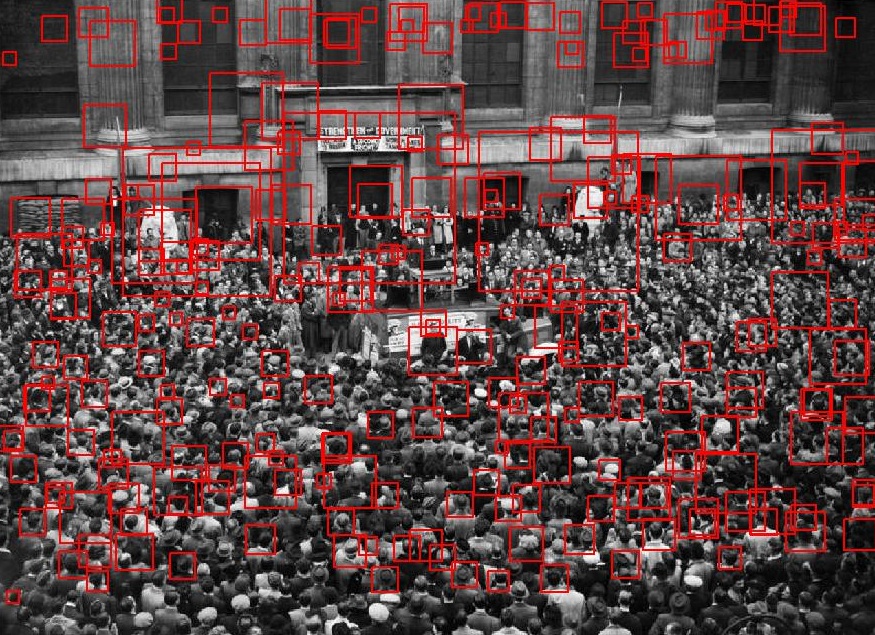}\\
\end{tabular}
\end{center}
\caption{Some head detection results.}
\label{fig:head}
\end{figure}

We see that there are many false positives and negatives in the detection results (Figure~\ref{fig:head}). However, we perform a lot better for nearby/larger heads. Since the texture analysis methods are crowd-blind and work well mostly for very dense crowds, we need SIFT-based analysis and head detections for adding robustness to the system such that we can perform accurately in relatively low-density environments too. 

Each detection is accompanied by the scale and confidence associated with it. For each cell we return the number of detection, $\eta_{head}$, and means and variances of the scales and confidences, 
to be used in the next step. 

\hspace{0.5cm}
\subsubsection{\textbf{Total count in the cell}}
\label{subsec:total}
We densely sample cells from the training images and obtain counts and other features from all the above methods. We then use the annotations to train an $\epsilon$-SVR with the counts and other features from above as inputs and the final estimate of the count as output. This SVR combines the information obtained from the five different sources to give an estimate of the patch count.
\\
\\
The total person count of the image is finally obtained by summing the counts obtained from all cells in the grid. Here we are assuming that the cell counts are independent. This is a reasonable assumption because we are dealing with widely varying viewpoints, perspective effects and crowd densities. We believe that putting neighbourhood constraints, as done in~\cite{Idrees2013}, limits the efficacy of multi-source count estimation to images with mostly uniform densities.

\section{Experiments}
\label{sec:exp&res}

\subsection{Dataset}
We first use the publicly available UCF crowd counting dataset to compare our results to past work. This dataset contains 50 images. These image contain 96 to 4633 people with an average of 1280 people per image. The authors of~\cite{Idrees2013} provide the ground truth dot-annotations with the images, i.e., each person is marked with a dot. There are 63974 annotations in the 50 images.

\begin{figure}
\begin{tabular}{cc}
\includegraphics[width=3.9cm,height=2.6cm]{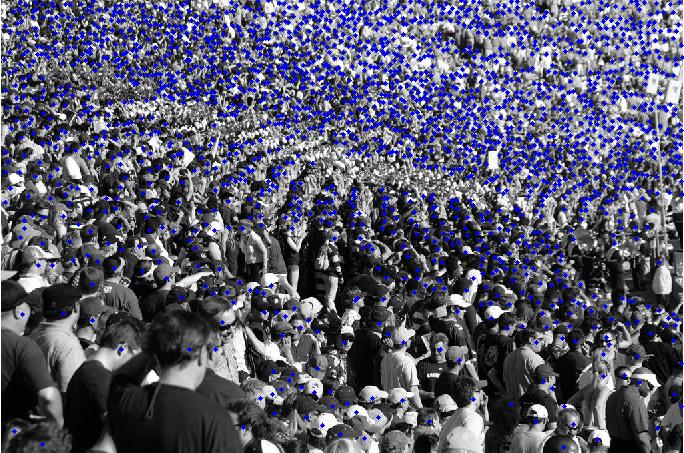}&
\includegraphics[width=3.9cm,height=2.6cm]{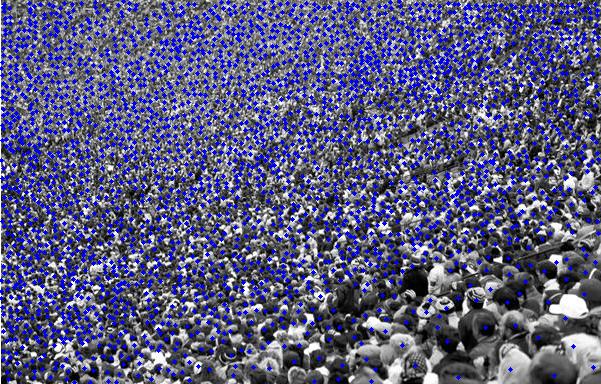}\\
\includegraphics[width=3.9cm,height=2.6cm]{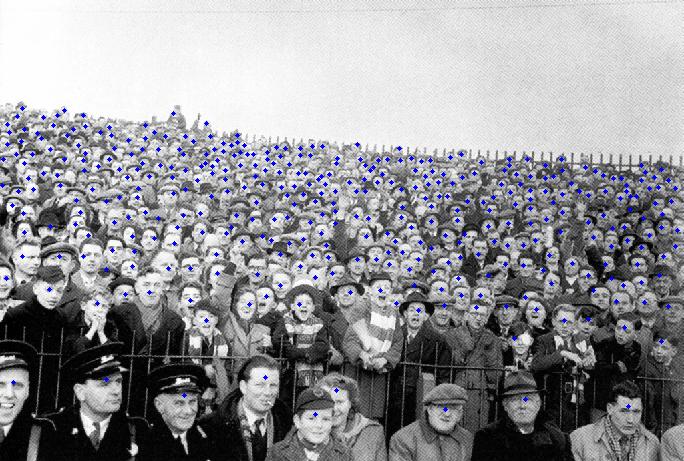}&
\includegraphics[width=3.9cm,height=2.6cm]{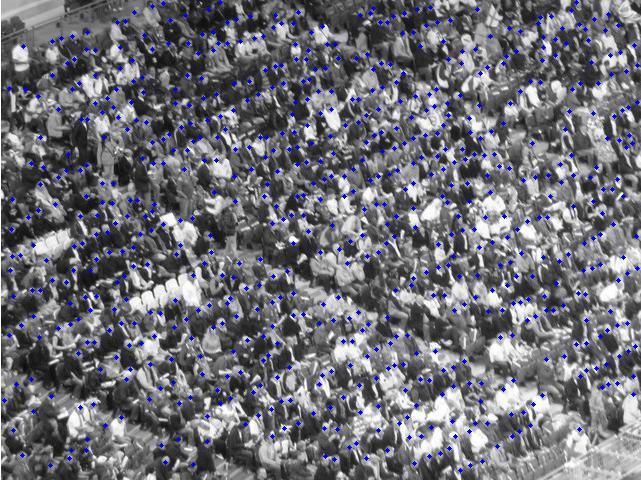}\\
\end{tabular}
\caption{Some annotated images from the dataset.}
\label{fig:ucfdtaset}
\end{figure}

We, then, collected 50 more images from Flickr and added annotations to extend the above dataset to 100 images. We included a huge variety of viewpoints and perspectives in these images. This was done to ensure that we have an estimate of the robustness of this system. A wider variety of perspective distortions was missing from the UCF dataset.

Finally we have 100 images with 87135 annotations containing, on average, 871 individuals per image with the number varying from 81 to 4633. 

\subsection{Evaluation metrics}
We use absolute error (AE) and normalised absolute error (NAE) for evaluating the performance.  We report the mean and deviations of both AE and NAE for both the UCF dataset and our extended dataset. 
\begin{eqnarray}
\mu_{AE} = \frac{1}{N}\sum_{i=1}^{N}|\eta_i - \hat{\eta_i}|\\
\mu_{NAE} = \frac{1}{N}\sum_{i=1}^{N}\frac{|\eta_i - \hat{\eta_i}|}{\eta_i}
\end{eqnarray}
where $\mu_{AE}$ and $\mu_{NAE}$ denote the mean of AE and NAE respectively, $\hat{\eta_i}$ is the estimated count, $\eta_i$ is the actual ground truth count, and $N$ is the number of cells/images.

Also, since we are dealing with small cells, we report the per-cell performances too.

\subsection{Evaluation}
We randomly divided the UCF dataset into groups of 10 and ran 5-fold cross validation. We compare the performance of our model with the models presented in~\cite{Idrees2013},~\cite{Rodriguez2011a} and~\cite{Lempitsky2010} in Table~\ref{tab:compare}. These methods are among the very few suited for this problem because most of the other methods either rely on videos or human detection, and cannot be used with the UCF dataset.

\begin{figure}
\begin{center}
\begin{tabular}{c}
\includegraphics[scale=0.38]{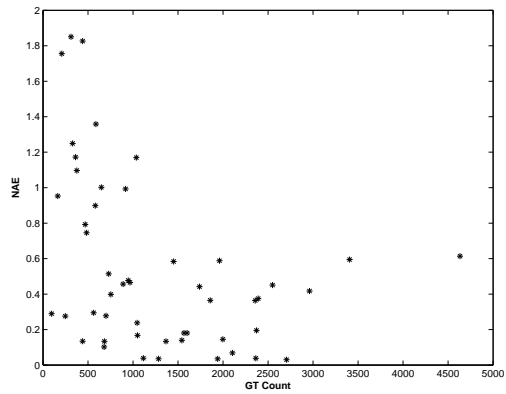}\\
(a)\\
\includegraphics[scale=0.38]{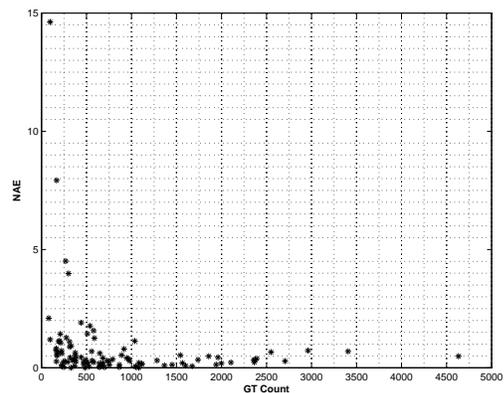}\\
(b)\\
\end{tabular}
\end{center}
\caption{Normalised Absolute Error (NAE) vs. the ground truth counts for (a) the UCF 50 images; and (b) the complete dataset.}
\label{fig:NAEvCount}
\end{figure}

The method presented by Rodriguez~\emph{et al.}~\cite{Rodriguez2011a} used head detections for counting while Lempitsky~\emph{et al.}~\cite{Lempitsky2010} used SIFT features to learn a regression function for counting. The authors of~\cite{Idrees2013} found that~\cite{Rodriguez2011a} performs best around counts of 1000, but as we move away on either side, its error increases. This because the estimated counts are fairly steady across the dataset and do not respond well to change in crowd density. It overestimates the counts in the low count images and underestimates in the high count images leading to high absolute error for both these cases. However,~\cite{Lempitsky2010} performs well at higher counts but poorly in terms of NAE for lower counts. The MESA-distance in~\cite{Lempitsky2010} is designed to minimise the maximum AE across image during training. Images with higher counts tend to have higher AE, and thus, the algorithm focusses mainly on these images. The model gets biased towards high density images but performs poorly for low density ones. From Figure~\ref{fig:NAEvCount}, we see that the proposed method too performs poorly for some low-count images. However, the method performs quite well in the middle and high count range ($>$1000 individuals per image).

We observe that, unlike our method, the other three methods perform poorly for the most high density images. All of them tend to underestimate the counts in that range. The first thing to note that most images in this category are very high-resolution. They have a very low chance of having individuals missed during annotation. Also, the per cell density increases super-linearly for this group, which is linear for other categories. Since, there are very few of such images, they could have been treated as outliers during training. Our method gives mostly average or better performances on such images. It relies heavily on the texture information from the images (Fourier analysis, GLCM features, wavelet features) to estimate the counts. So it performs well on the higher density crowds, since the texture approximation is best applicable to such crowds.

We also evaluated the performance of our proposed method on the extended dataset containing 100 images. This dataset has much more diversity than the original dataset because it has crowds present in varying densities and visible from various viewpoints. This is useful for testing the robustness of the algorithm. For this case, we divided the dataset into sets of 25 and ran 4-fold cross validation. The final per-patch and per-image results are shown in Table~\ref{tab:final100}. 

\begin{table}
\begin{center}
\rowcolors{1}{white}{light-gray}
\begin{tabular}{ccc}
\hline
\textbf{Method} & \textbf{AE} & \textbf{NAE} \\ \hline
Rodriguez \emph{et al.} & 655.7 $\pm$ 697.8 & 0.706 $\pm$ 1.02 \\ \hline
Lempitsky \emph{et al.} & 493.4 $\pm$ 487.1 & 0.612 $\pm$ 0.916 \\ \hline
\textbf{Proposed} & 514.1 $\pm$ 526.4 & 0.542 $\pm$ 0.484 \\ \hline
Idrees \emph{et al.} & 419.5 $\pm$ 541.6 & 0.313 $\pm$ 0.271 \\ \hline
\end{tabular}
\end{center}
\caption{Quantitative comparison of the proposed method with Rodriguez \emph{et al.}~\cite{Rodriguez2011a}, Lempitsky \emph{et al.}~\cite{Lempitsky2010}, and Idrees \emph{et al.}~\cite{Idrees2013} using the means and standard deviations of Absolute Error (AE) and Normalised Absolute Error (NAE) for the UCF crowd counting dataset. The proposed algorithm out-performs \cite{Rodriguez2011a} and \cite{Lempitsky2010}, but is outperformed by a much more computationally expensive model from \cite{Idrees2013}.}
\label{tab:compare}
\end{table} 

\begin{table}
\begin{center}
\begin{tabular}{ccc}
\hline
 & \textbf{AE} & \textbf{NAE} \\ \hline
\textbf{Per-patch} & 9.5 $\pm$ 14.682 & - \\ \hline
\textbf{Per-image} & 377.7 $\pm$ 480.8 & 0.666 $\pm$ 1.123 \\ \hline
\end{tabular}
\end{center}
\caption{Per-patch and per-image results for the complete dataset of 100 images.}
\label{tab:final100}
\end{table}

\begin{figure}
\begin{center}
\begin{tabular}{cc}
\includegraphics[width=3.8cm,height=2.4cm]{53.jpg}&
\includegraphics[width=3.8cm,height=2.4cm]{89.jpg}\\
Est: 641.5, GT: 644 & Est: 1668.1, GT: 1674\\
\includegraphics[width=3.8cm,height=2.4cm]{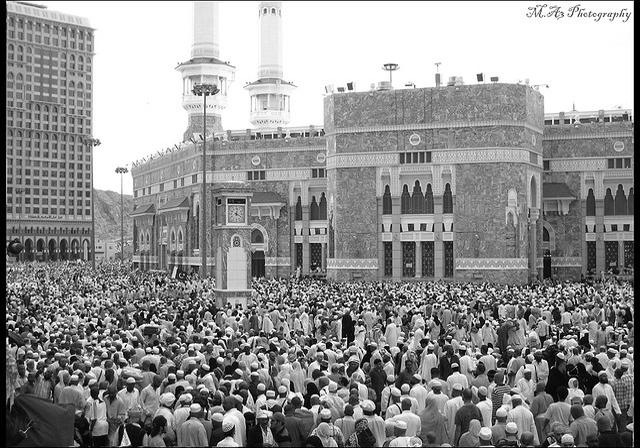}&
\includegraphics[width=3.8cm,height=2.4cm]{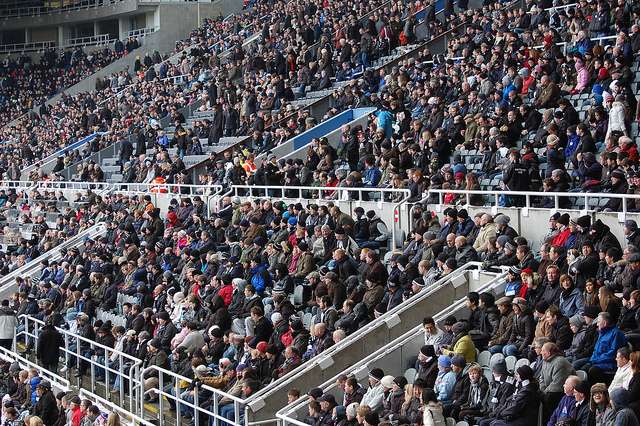}\\
Est: 708, GT: 700 & Est: 852.6, GT: 864\\
\includegraphics[width=3.8cm,height=2.4cm]{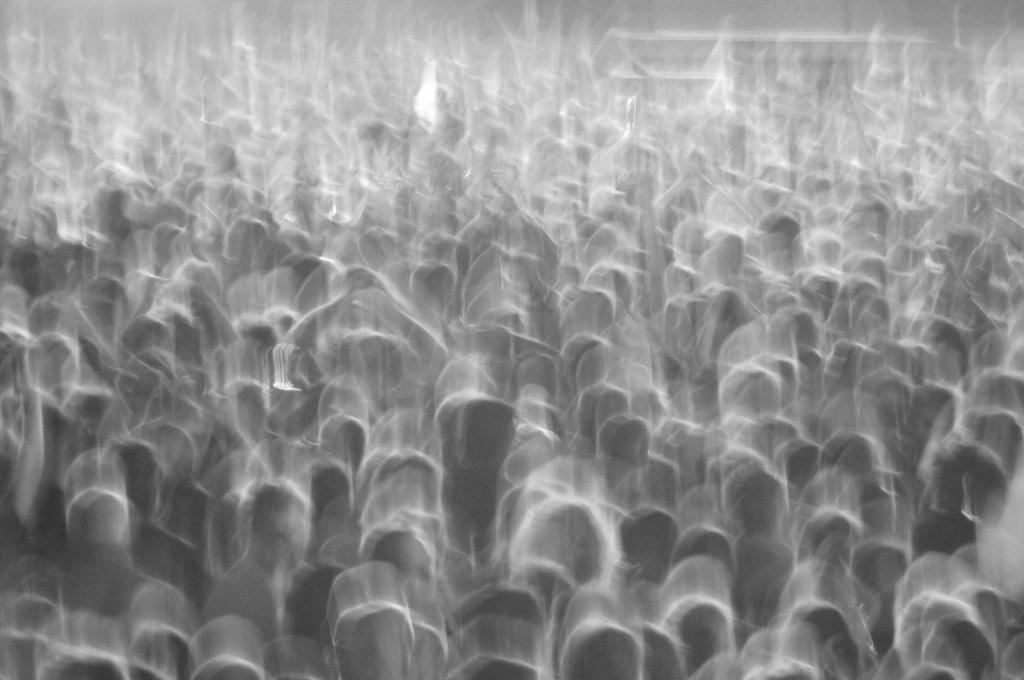}&
\includegraphics[width=3.8cm,height=2.4cm]{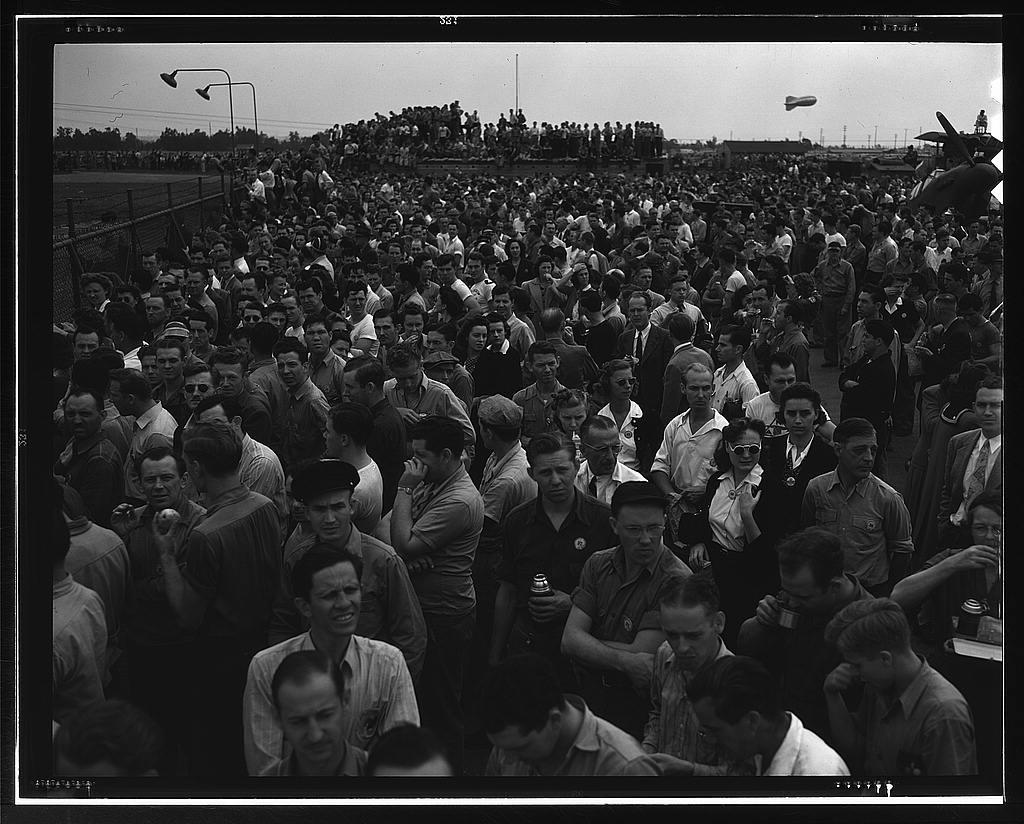}\\
Est: 84, GT: 96 & Est: 602.8, GT: 581\\
\end{tabular}
\end{center}
\caption{Estimated counts for some images with the lowest absolute errors (AE).}
\label{fig:least_abs_err}
\end{figure}

\begin{figure}
\begin{center}
\begin{tabular}{cc}
\includegraphics[width=3.8cm,height=2.4cm]{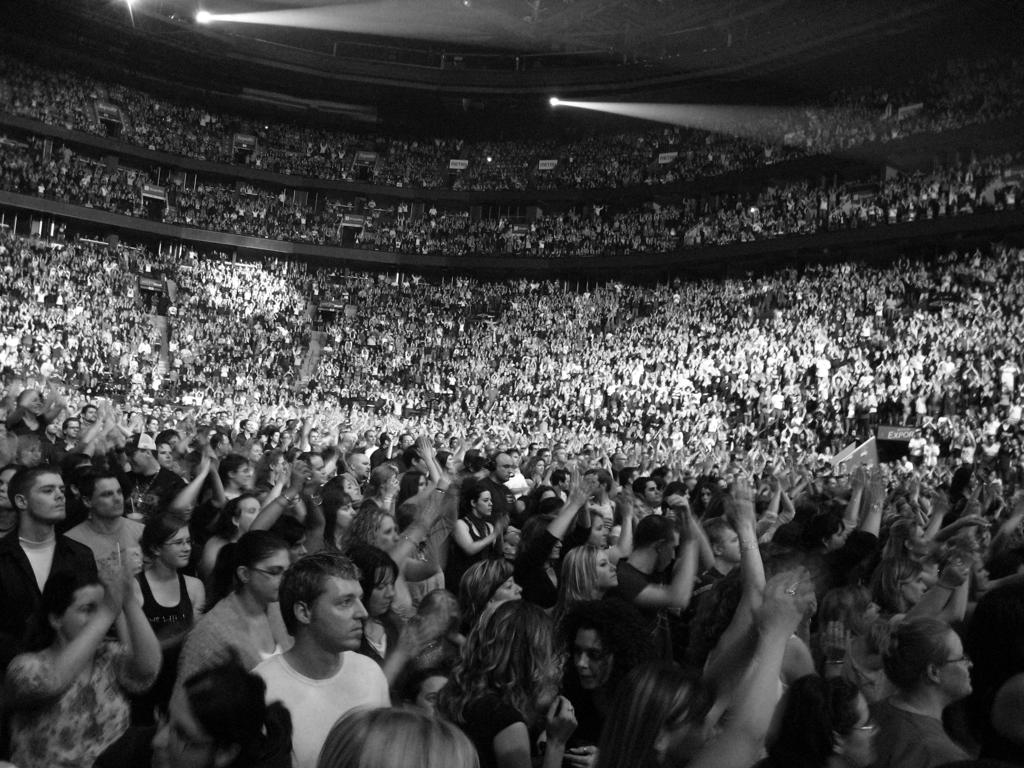}&
\includegraphics[width=3.8cm,height=2.4cm]{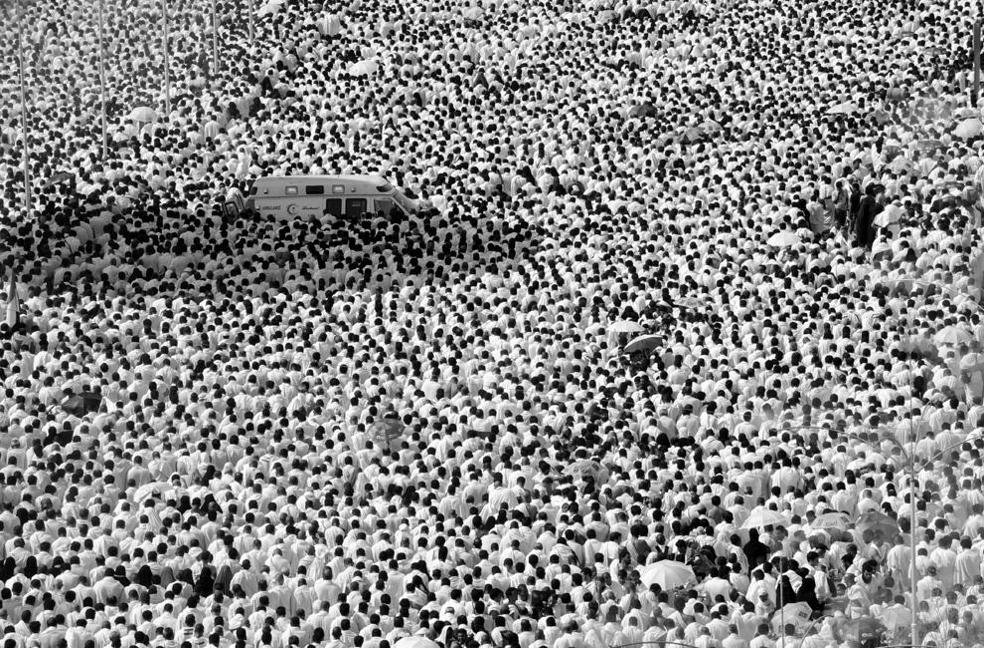}\\
Est: 1012.3, GT: 3406 & Est: 1557.9, GT: 4633\\
\includegraphics[width=3.8cm,height=2.4cm]{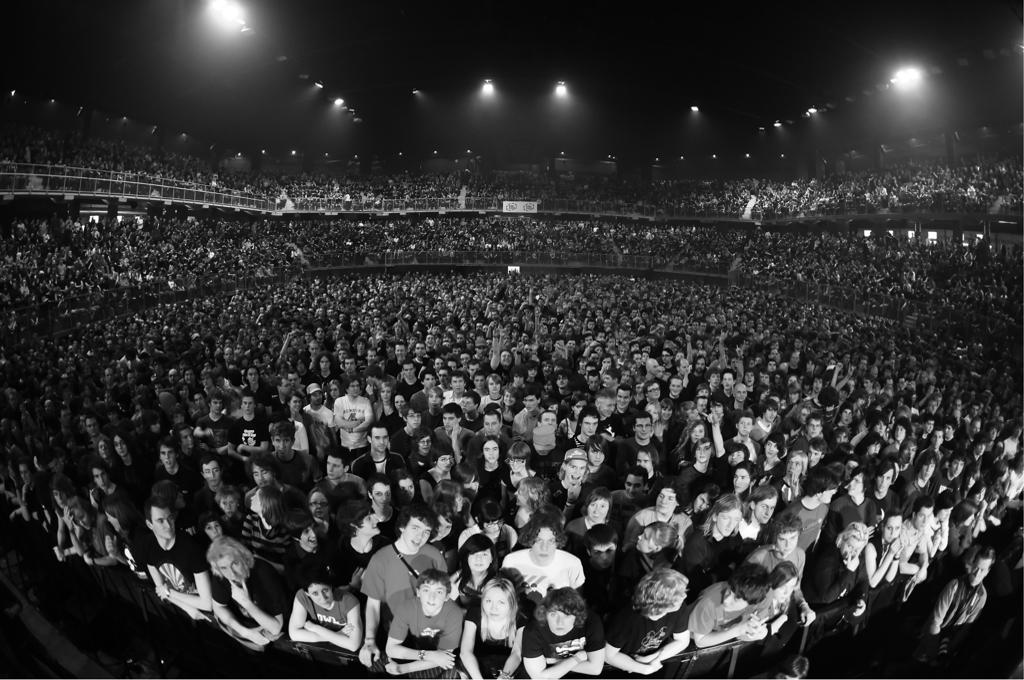}&
\includegraphics[width=3.8cm,height=2.4cm]{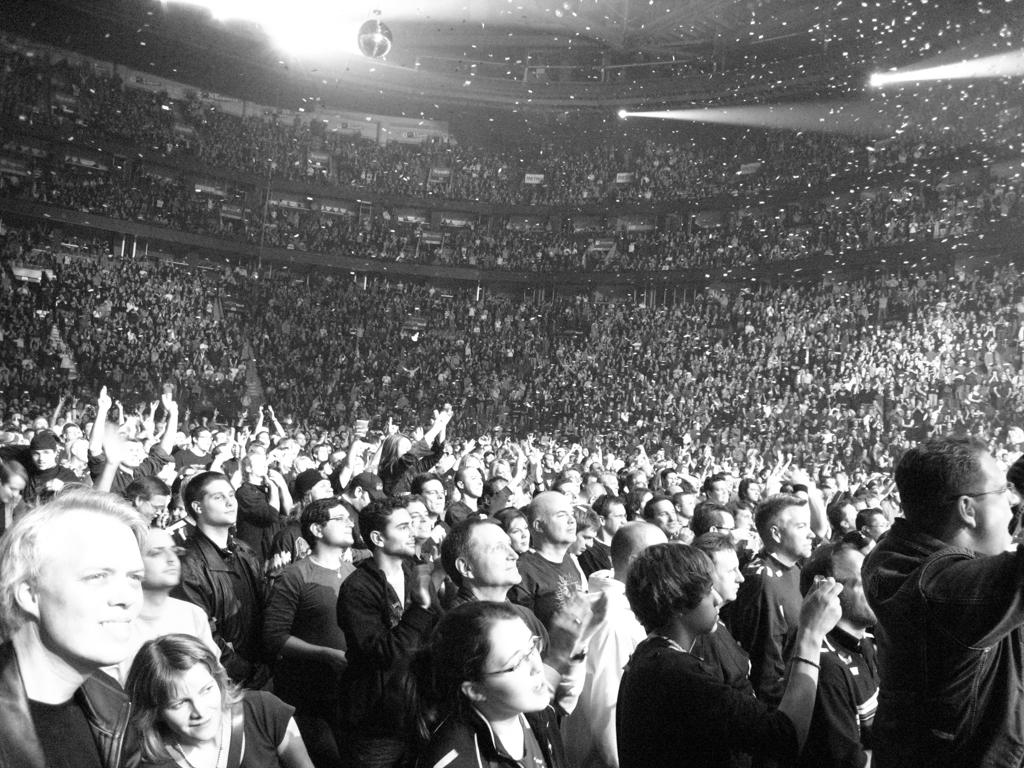}\\
Est: 958.7, GT: 2961 & Est: 1025.3, GT: 2550\\
\includegraphics[width=3.8cm,height=2.4cm]{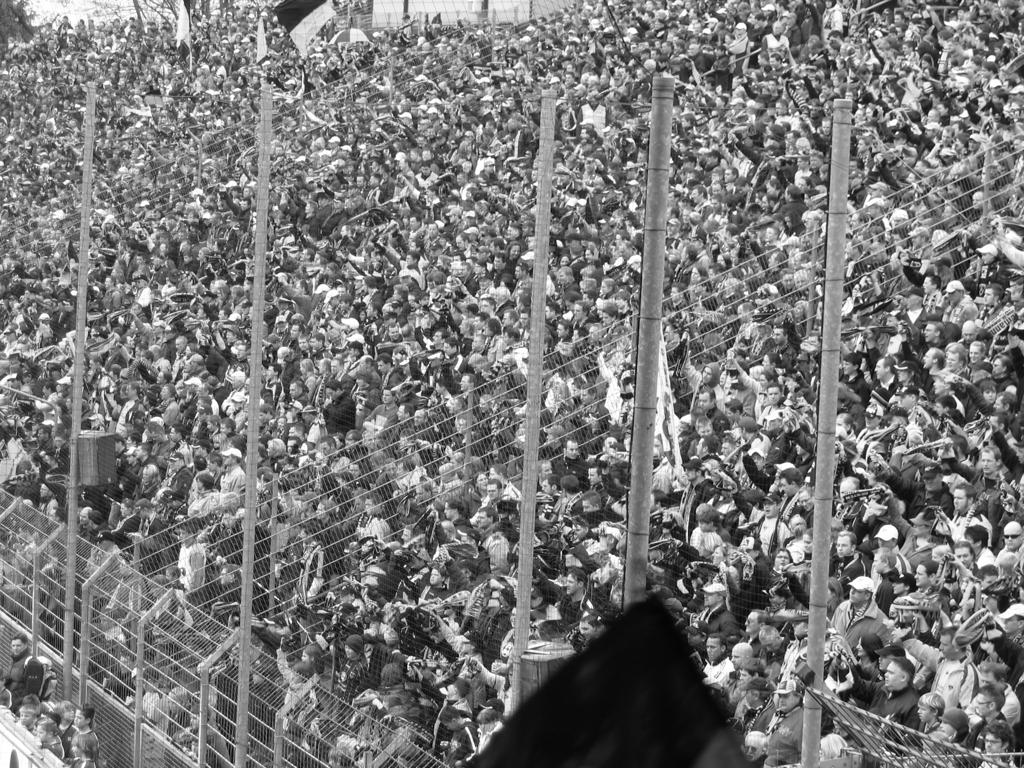}&
\includegraphics[width=3.8cm,height=2.4cm]{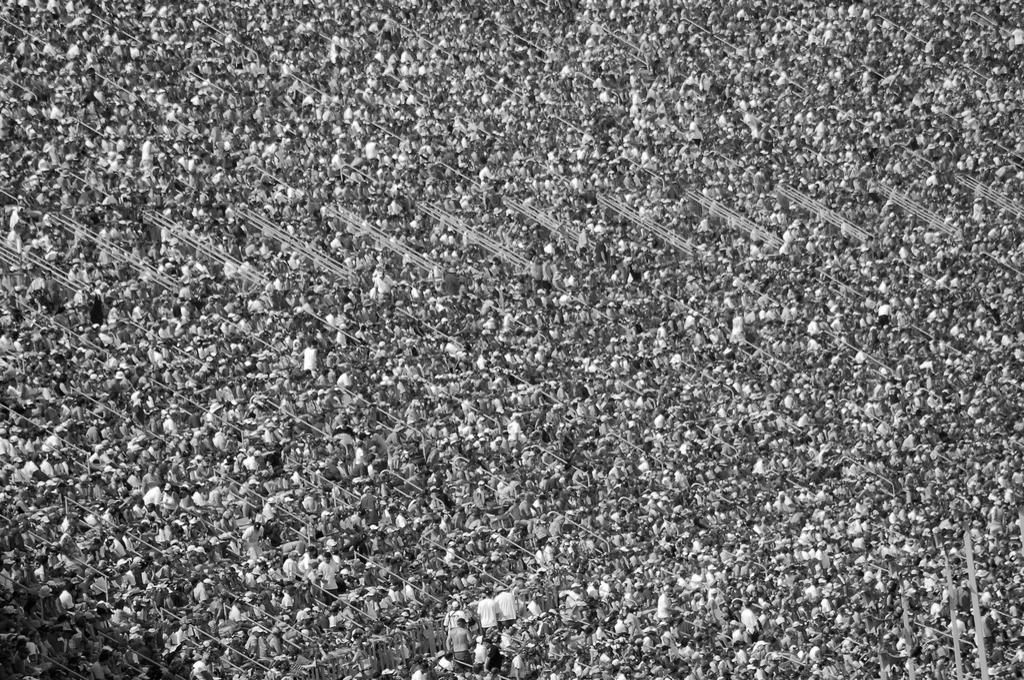}\\
Est: 2165.7, GT: 1037 & Est: 3445, GT: 2358\\
\end{tabular}
\end{center}
\caption{Estimated counts for some images with the highest absolute errors (AE).}
\label{fig:most_abs_err}
\end{figure}

Table~\ref{tab:final100} gives the performance of the algorithm on the final dataset in terms of mean absolute error and mean normalised absolute error at both the patch level and image level. We obtain a mean absolute error of 377.7 with a standard deviation of 480.8 and a mean normalised error of 0.666 with standard deviation 1.123. The reason of a seemingly higher NAE is evident from Figure~\ref{fig:NAEvCount}. There are a very small number of images in the low crowd-density category (below 500 individuals per image range) which drive the mean NAE up. Our method does not work quite well for some low density images. Figures~\ref{fig:least_abs_err} and~\ref{fig:most_abs_err} show the images with the lowest and highest absolute errors respectively.

We observe that most of the images for which we get high absolute errors are very high density crowds. These images mostly contain extreme perspective variations. Also, some of the images have lens distortions which may be a reason for poor estimates. We also note that the NAE is very high for some of the images in the low density region. We believe that texture methods do not perform very well for such images. We are using head detections and interest points analysis as parts of our system. Further research in this area could focus on finding ways to pre-determine the density of crowds in different image regions so that these methods (head detections and interest-point analysis) could be given more importance for those regions. We also note that only a very few images are driving the average absolute error and NAE up. Removing just the worst 10\% performing images from the final dataset and considering the rest 90 images reduces the absolute error to 256.3 $\pm$ 217.7 and the NAE to a very low 0.407 $\pm$ 0.328.

\begin{figure}
\begin{center}
\includegraphics[scale=0.45]{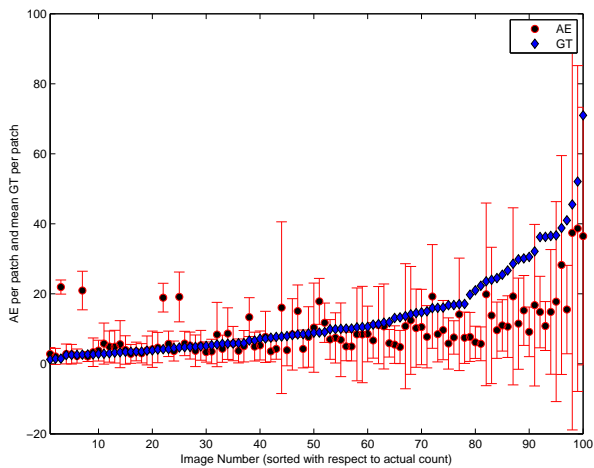}
\end{center}
\caption{Analysis of patch estimates in terms of absolute error per patch. The image numbers have been sorted with respect to the actual counts. Black dots are the mean absolute errors, red bars represent the standard deviations and blue diamonds are the ground truths.}
\label{fig:perpatchanalysis}
\end{figure}

Figure~\ref{fig:perpatchanalysis} shows the per patch performance of the algorithm. Black dots are the mean absolute errors per patch, red bars represent the standard deviations, and blue diamonds are the actual average number of individuals per patch. We observe that, for higher density crowds, the mean absolute error per patch increase with increase in actual count. The absolute error per patch is almost constant, and very small, till around image 90, i.e., for images with counts less than about 2000. This is a demonstration of the efficacy of algorithm presented.

%
%

\section{Conclusion}
\label{conclusion}
We considered a method for estimating the number of people in extremely dense crowds from still images. The counting problem at this scale has barely been tackled before. We presented a method that uses information from multiple sources to estimate the count in an image. We used head detections, interest points based counting and texture analysis methods (Fourier analysis, GLCM features and wavelet analysis) as the different sources of information. Each of these constituent parts gives an independent estimate of the count, along with confidences and other features, which are then fused to give a final estimate. We present results of extensive tests and experiments we performed. We also introduced a new dataset of still images along with annotations which can complement the existing UCF dataset. The results are very promising and, since the model is extremely simple, it can be applied for real-time counting in critical areas like pilgrimage sites and other areas where which present a danger of stampedes.


%




\ifCLASSOPTIONcaptionsoff
  \newpage
\fi



%

\newpage
\bibliographystyle{IEEEtran}
\bibliography{IEEEabrv,thesis}

\end{document}